\documentclass[conference]{IEEEtran}
\IEEEoverridecommandlockouts

\usepackage{amsmath,amssymb,amsfonts}
\usepackage{booktabs}
\usepackage{multirow}
\usepackage{colortbl}
\usepackage{array}
\usepackage{multicol} 
\usepackage{hyperref}

\usepackage{algorithm}
\usepackage{algpseudocode}

\usepackage{graphicx}
\usepackage{tikz}
\usepackage{caption}
\usepackage{subcaption}
\usepackage{xcolor}
\usepackage{colortbl} 
\usepackage{orcidlink}

\def\BibTeX{{\rm B\kern-.05em{\sc i\kern-.025em b}\kern-.08em
    T\kern-.1667em\lower.7ex\hbox{E}\kern-.125emX}}
\begin{document}

\title{Guided and Unguided Conditional Diffusion Mechanisms for Structured and Semantically-Aware 3D Point Cloud Generation\\
\thanks{This work was supported in part by the National Science Foundation under Grant No. OIA-2148788.}
}

\author{
\IEEEauthorblockN{Gunner Stone\orcidlink{0000-0002-5182-1365}$^*$, Sushmita Sarker\orcidlink{0000-0003-0390-3099}$^*$, and Alireza Tavakkoli\orcidlink{0000-0001-9460-1269}\thanks{$^*$ Equal contribution}}
\IEEEauthorblockA{\textit{Department of Computer Science and Engineering, University of Nevada, Reno,}
USA}
}

\maketitle

\begin{figure*}[ht]
    \centering
    \includegraphics[width=\textwidth]{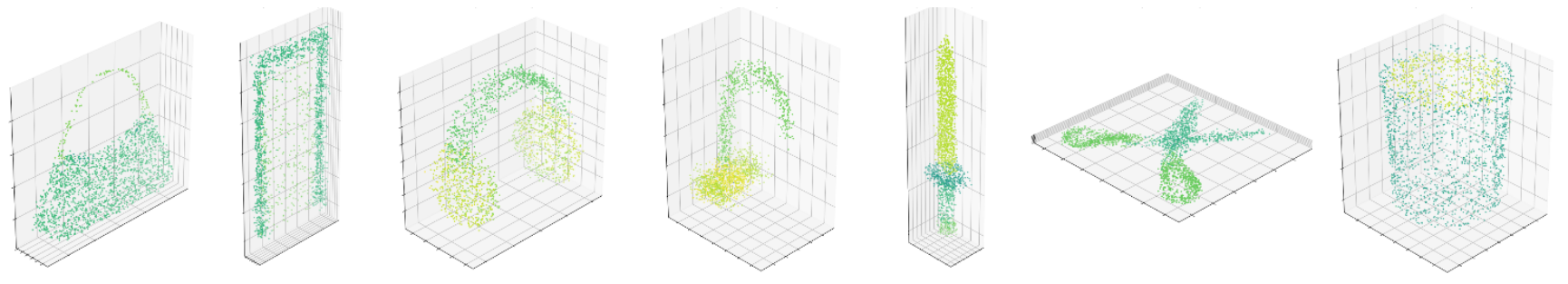}
    \caption{Examples of 3D point cloud generation with part-wise semantic encoding. Our method generates semantically annotated point clouds for various objects: bag, door, earphones, faucet, knife, scissors, and trashcan. The color-coded points represent different parts of each object, demonstrating the ability of our approach to produce detailed and part-specific structures. As opposed to conventional point cloud generative models that only perform object-level generation, our guided and unguided point-based diffusion processes enable accurate and semantically meaningful generation of complex 3D shapes.}
    \label{fig:teaser}
\end{figure*}

\begin{abstract}
Generating realistic 3D point clouds is a fundamental problem in computer vision with applications in remote sensing, robotics, and digital object modeling. Existing generative approaches primarily capture geometry, and when semantics are considered, they are typically imposed post hoc through external segmentation or clustering rather than integrated into the generative process itself. We propose a diffusion-based framework that embeds per-point semantic conditioning directly within generation. Each point is associated with a conditional variable corresponding to its semantic label, which guides the diffusion dynamics and enables the joint synthesis of geometry and semantics. This design produces point clouds that are both structurally coherent and segmentation-aware, with object parts explicitly represented during synthesis. Through a comparative analysis of guided and unguided diffusion processes, we demonstrate the significant impact of conditional variables on diffusion dynamics and generation quality. Extensive experiments validate the efficacy of our approach, producing detailed and accurate 3D point clouds tailored to specific parts and features. All code can be found on this project's \href{https://github.com/sushmitaSARKER/Conditional-Diffusion-Model-for-Semantically-Aware-3D-Point-Cloud-Generation.git}{Github}.
\end{abstract}

\section{Introduction}
\label{sec:intro}

Generative models are a cornerstone of unsupervised learning, supporting tasks such as shape completion, upsampling, and 3D synthesis. Extending these capabilities to point clouds is especially important for applications in remote sensing, robotics, and digital object modeling, where accurate 3D representations are critical. However, creating realistic point clouds remains difficult. While discriminative methods for segmentation and classification have advanced significantly~\cite{qi2017pointnet,qi2017pointnet++,wang2019dynamic}, progress in generative modeling is still an ongoing problem, particularly for complex 3D structures.

Conventional generative models such as VAEs~\cite{kingma2013auto} and GANs~\cite{goodfellow2014generative} struggle with the irregular and sparse structure of point cloud data. While highly effective for 2D image generation, they do not transfer well to 3D domains that lack a regular grid. Diffusion models, in contrast, show promise in 3D by representing point clouds as particle systems and applying stochastic dynamics to iteratively refine their structure.

In this work, we introduce a diffusion-based framework for 3D shape generation that directly incorporates semantic understanding into the generative process. Unlike prior diffusion models for point clouds, which focus solely on producing geometrically plausible shapes~\cite{luo2021diffusion,li2022diffusionpointlabel}, our approach integrates per-point semantic labels as conditional variables. These labels are not added post hoc but actively guide the diffusion dynamics at every step. As a result, the model generates point clouds that are both structurally coherent and semantically meaningful, with object parts explicitly represented during synthesis. This capability enables fine-grained control and supports applications that require detailed recognition of individual components.

Building on prior work that adapts diffusion processes from non-equilibrium thermodynamics to point cloud generation~\cite{luo2021diffusion}, we extend this foundation by embedding per-point conditional variables directly into the diffusion process. These variables correspond to semantic part labels and are jointly learned with the model parameters. By conditioning generation on semantics at the level of individual points, the model produces point clouds that are both geometrically faithful and segmentation-aware.

To evaluate the role of these conditional variables, we compare guided and unguided diffusion processes. In the guided setting, semantic variables remain fixed and consistently steer generation toward coherent object parts. In the unguided setting, these variables are perturbed stochastically, which weakens semantic consistency and alters diffusion dynamics. This comparison demonstrates how explicit semantic conditioning improves both structural fidelity and part-level interpretability in 3D point cloud synthesis.

We summarize our contributions as follows:

\begin{itemize}
\item We introduce a diffusion-based generative framework for point clouds that incorporates per-point semantic conditioning, enabling joint shape and part-level synthesis.
\item We conduct a systematic study of guided and unguided diffusion, analyzing how per-point conditioning influences generation quality, segmentation fidelity, and common failure modes across different granularities.
\end{itemize}

\section{Related Works}

\textbf{Point Cloud Generative Models:}
Previous research in point cloud generation has explored autoencoding \cite{gadelha2018multiresolution, yang2018foldingnet}, single-view reconstruction \cite{groueix2018papier, fan2017point, klokov2020discrete, kurenkov2018deformnet}, and adversarial generation \cite{valsesia2018learning, zamorski2020adversarial, shu20193d}. These methods often rely on heuristic loss functions like Chamfer Distance (CD) and Earth Mover's Distance (EMD). A shift in perspective treats 3D point clouds as probabilistic distributions, leading to innovations like autoregressive generation by Sun et al.\cite{sun2020pointgrow}, which requires ordering of point clouds. Probabilistic approaches also manifest in GAN-based and flow-based models\cite{klokov2020discrete, achlioptas2018learning, yang2019pointflow, li2018point, kim2020softflow}, where notable examples include PointFlow~\cite{yang2019pointflow}, which utilizes normalizing flow techniques, and Discrete PointFlow, which adopts discrete normalizing flow with affine coupling layers~\cite{papamakarios2021normalizing}. Diverging from these, Shape Gradient Fields~\cite{cai2020learning} learn a gradient field for sampling point clouds through Langevin dynamics. 

Despite these advances, existing methods face inherent limitations. GANs often suffer from training instability and mode collapse, while autoregressive models impose an unnatural fixed order on the unordered nature of point clouds. VAEs and flow-based models struggle with the complex distributions of 3D data, often requiring extensive tuning. Given these challenges, diffusion models offer a promising alternative by treating point clouds as dynamic particle systems, naturally accommodating their irregular structure and leveraging stochastic dynamics to refine samples over time.

\textbf{Energy-Based and Denoising Diffusion Models:}
Two distinct strategies for iterative generation are Energy-Based Models (EBMs) \cite{nijkamp2019learning, du2019implicit, lecun2006tutorial} and Denoising Diffusion Models \cite{sohl2015deep, song2020denoising, ho2020denoising}. EBMs define an energy landscape where local minima correspond to valid data samples, typically generated via Langevin dynamics. Denoising diffusion models, by contrast, progressively denoise Gaussian noise to recover the target data distribution, a process akin to score matching in EBMs. 

Building on these principles, Luo et al.~\cite{luo2021diffusion} introduced a diffusion framework tailored for 3D point clouds, treating them as particle systems and demonstrating how stochastic dynamics can capture complex geometric structure. Subsequent work such as DiffusionPointLabel \cite{li2022diffusionpointlabel} extended this idea by producing semantically annotated point clouds. While a step in the right direction, their method relies on a separate feature interpreter to extract point-wise labels from a pre-trained diffusion model, leaving semantics as an auxiliary prediction rather than an integral part of generation.

\section{Background}

Traditional diffusion probabilistic models, which are designed for unconditional generation, produce samples from random noise to represent the data distribution. Instead, this work is based on conditional generation, specifically creating 3D point clouds based on shape latents. In this approach based on work done by Luo\cite{luo2021diffusion}, a shape latent is incorporated as a condition at each step in the reverse Markov chain, fundamentally altering training and sampling strategies.

\subsection{Diffusion Probabilistic Models for Point Clouds}

Diffusion probabilistic models generate data by reversing a Markovian noising process \cite{sohl2015deep}. For point clouds, an input $\mathbf{X}_0=\{x_i\}_{i=1}^N$ is gradually perturbed with Gaussian noise over $T$ steps, producing $\mathbf{X}_T$ that approaches an isotropic Gaussian. The forward kernel is
\begin{equation}
q(x_t|x_{t-1}) = \mathcal{N}(\sqrt{1-\beta_t}\,x_{t-1}, \beta_t \mathbf{I}),
\end{equation}
where $\{\beta_t\}$ defines the variance schedule. 

The reverse diffusion process reconstructs the point cloud by traversing a reverse Markov chain from a noise distribution, $p(x_T)=\mathcal{N}(0,I)$, using a neural network, $\mu_\theta$:

\begin{equation}
p_\theta(x_{t-1}|x_t,z) = \mathcal{N}(\mu_\theta(x_t,t,z), \beta_t \mathbf{I}),
\end{equation}
The model $\mu_{\theta}$ estimates the mean at each time step, conditioned on the current state and the shape latent $z$.

\subsection{Conditional Model Architecture and Diffusion Process}

Our conditional diffusion model has two components: an encoder and a decoder. The encoder maps the input point cloud \(x_0\) to a latent vector \(z=E(x_0)\), which summarizes its structure. A KL regularization term encourages \(z\) to follow a Gaussian prior, allowing new samples to be drawn for generation, similar to a variational autoencoder.

The decoder predicts the noise injected at step \(t\):
\begin{equation}
e_\theta = D(z, t, x_t).
\end{equation}
During denoising, this estimate is subtracted to recover the previous state, progressively refining noisy inputs back into coherent point clouds.
\begin{equation}
x_{t-1,i} \approx x_{t,i} - e_{\theta,i},
\end{equation}

This work explores two distinct diffusion processes: guided and unguided, differentiated by their handling of point-level class information.

\begin{figure*}[!ht]
    \centering
    \resizebox{0.9\textwidth}{!}{%
        \begin{minipage}{\textwidth}
            \begin{subfigure}[t]{0.49\textwidth}
                \centering
                \includegraphics[width=\textwidth]{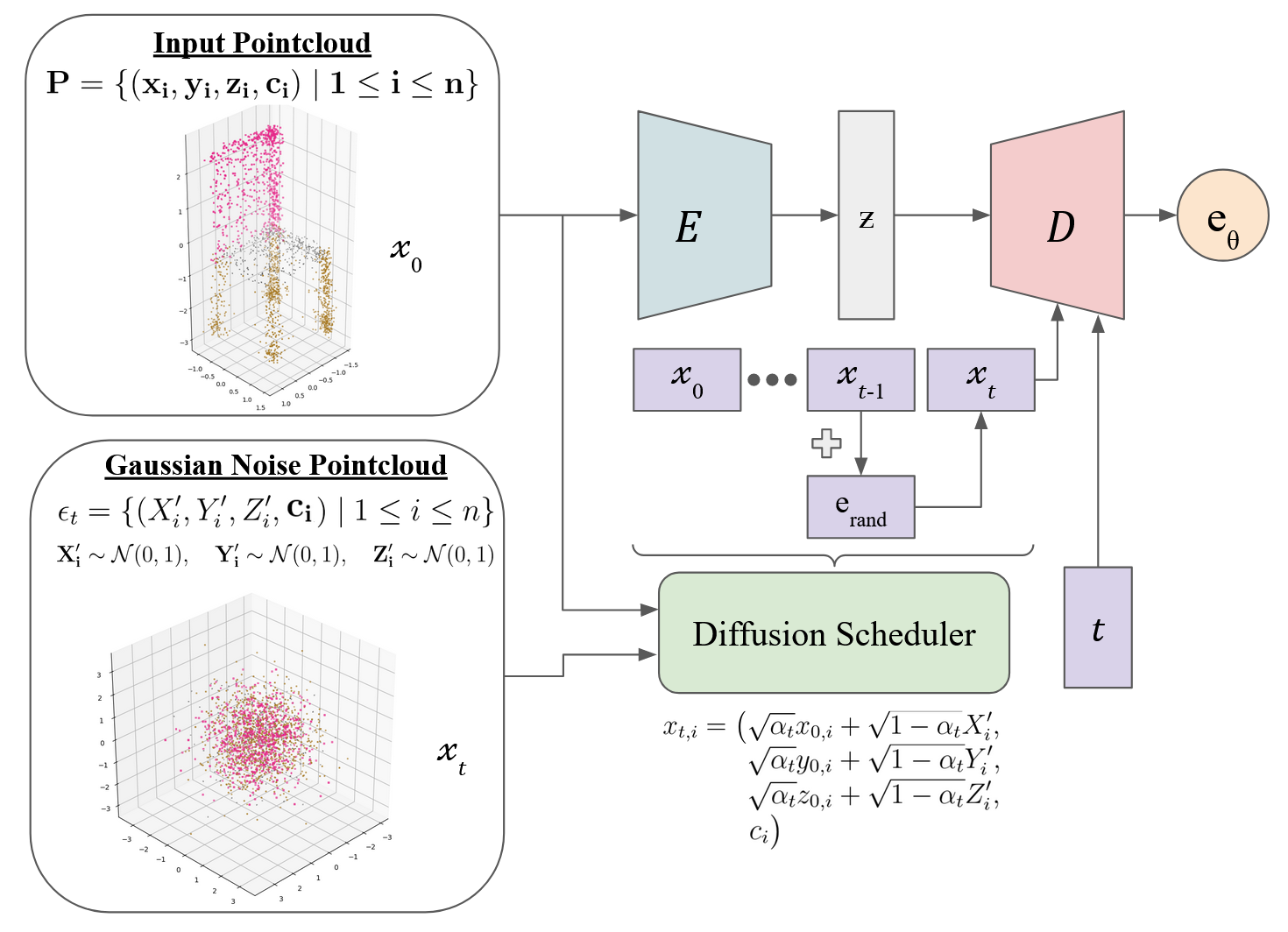}
                \caption{Guided Diffusion Model. Gaussian noise is added to spatial coordinates 
        \((X'_i, Y'_i, Z'_i \sim \mathcal{N}(0,1))\) at each timestep, while class labels 
        \(c_i\) remain fixed.}
                \label{fig:guided_archi}
            \end{subfigure}%
            \hfill
            \begin{subfigure}[t]{0.49\textwidth}
                \centering
                \includegraphics[width=\textwidth]{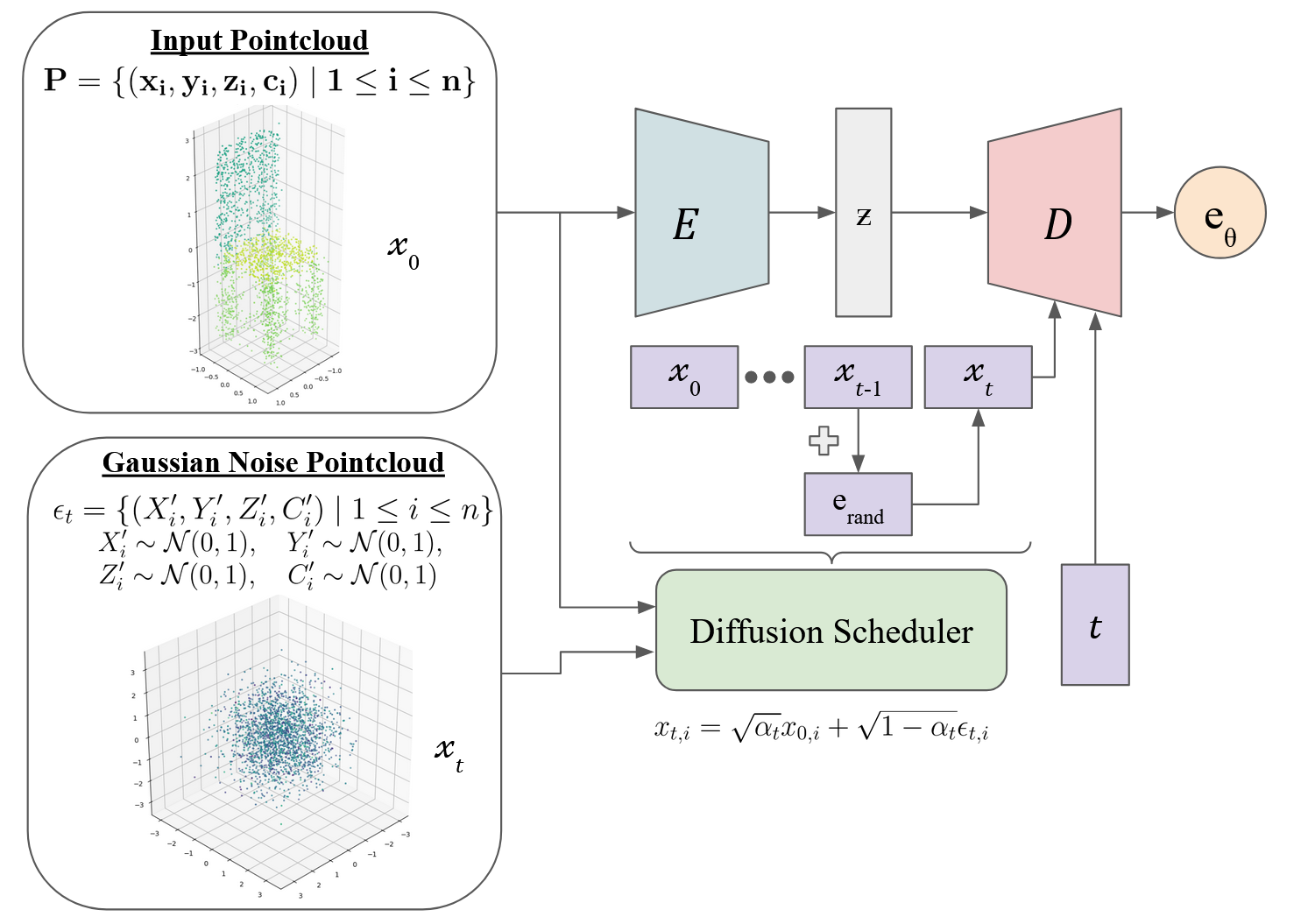}
                \caption{Unguided Diffusion Model. Gaussian noise is added to both spatial 
        coordinates and labels \((X'_i, Y'_i, Z'_i, C'_i \sim \mathcal{N}(0,1))\), 
        weakening semantic consistency.}
                \label{fig:archi_unguided}
            \end{subfigure}
        \end{minipage}%
    }
    \caption{Comparison of Guided and Unguided Diffusion Model Architectures for 3D Point Cloud Generation.}
    \label{fig:diffusion_models}
\end{figure*}

\subsection{Guided Point-Based Diffusion Process}

The guided variant incorporates per-point labels by noising only spatial coordinates while keeping class information fixed \((C_i = \mathbf{P}_{c_i})\)(Fig.~\ref{fig:guided_archi}). Each input point cloud is represented as
\(\mathbf{P}=\{(x_i,y_i,z_i,c_i)\}_{i=1}^n\), where $(x,y,z)$ are coordinates and $c$ is a semantic label. The encoder maps $\mathbf{P}$ to a latent $z$, which conditions the decoder during denoising.

The Diffusion Scheduler controls the transformation of the point cloud by linearly combining the original data \(x_0\) and the noise components \( \epsilon_t \) at each timestep, formulated as:

\begin{equation}
\begin{aligned}
x_{t, i} = (&\sqrt{\alpha_t} \mathbf{P}_{x_{0,i}} + \sqrt{1 - \alpha_t} X'_i, \\
&\sqrt{\alpha_t} \mathbf{P}_{y_{0,i}} + \sqrt{1 - \alpha_t} Y'_i, \\
&\sqrt{\alpha_t} \mathbf{P}_{z_{0,i}} + \sqrt{1 - \alpha_t} Z'_i, \mathbf{P}_{c_i})
\end{aligned}
\end{equation}

This equation ensures that while the spatial components (\(x_0\)) are progressively transitioned towards the noise state, the class component (\(c\)) remains unchanged, emphasizing the preservation of categorical integrity throughout the diffusion process.

The scheduler adjusts noise levels across timesteps from \(x_0\) to \(x_t\). Here, \( \alpha_t \) dictates the balance between the original data and the added noise, facilitating the generation of intermediate states within the diffusion process. The role of the diffusion scheduler is to manage this balance throughout the diffusion timeline, effectively simulating the gradual transition of the point cloud from structured to randomized states.

The decoder predicts the injected noise $e_\theta=D(z,t,x_t)$, which is subtracted to recover the previous state:
\[
x_{t-1,i} \approx (x_{t,i}-e_{\theta,i}, \; c_i).
\]

\noindent\textbf{Loss Function:}
Training combines two terms. First, an MSE loss ensures accurate prediction of spatial noise, while the class component is penalized toward zero (since no label noise is added):
\begin{equation}
\text{MSE}=\tfrac{1}{n}\sum_{i=1}^n\big(\|e_{\theta,i,xyz}-e_{\text{rand},i}\|^2 + (e_{\theta,i,c})^2\big).
\end{equation}

This MSE formulation helps quantify the discrepancies between the predicted and actual noise vectors specifically applied to transition between successive timesteps. This targeted optimization function aims to refine the guided diffusion process and ensure more accurate reconstructions of 3D point clouds.

Second, a Per-Class Chamfer Distance (CD) compares reconstructions within each semantic class:
\begin{align}
\text{CD}_c(P_c, Q_c) &= \frac{1}{|P_c|} \sum_{p \in P_c} \min_{q \in Q_c} \|p - q\|^2 \label{eq:CD_c} \\
&\quad + \frac{1}{|Q_c|} \sum_{q in Q_c} \min_{p \in P_c} \|p - q\|^2 \notag \\
\text{CD}_{\text{average}} &= \frac{1}{C} \sum_{c} \text{CD}_c(P_c, Q_c)
\end{align}

Unlike a global CD metric, which evaluates distances between all points regardless of their semantic meaning, the Per-Class CD loss restricts distance calculations to pairs of points within the same semantic class. This constraint is crucial as it prevents the model from achieving artificially low error rates by aligning points from different classes that are spatially close but semantically distinct.

Overall, the guided loss function can be summarized as a combination of the MSE and the Per-Class CD losses. The total loss function is formulated as:
\begin{equation}
\mathcal{L}_{\text{guided}}=\text{MSE}+\text{CD}_{\text{avg}} \label{eq:combined_loss}
\end{equation}
This encourages precise denoising while preserving semantic alignment across parts.

\subsection{Unguided Point-Based Diffusion Process}

In the unguided variant (Fig.~\ref{fig:archi_unguided}), both spatial coordinates and class labels are noised. Each noise vector is drawn from a standard Gaussian, 
\(\epsilon_{t,i}=(X'_i,Y'_i,Z'_i,C'_i)\sim \mathcal{N}(0,I).\)  
The forward step is
\begin{equation}
x_{t,i}=\sqrt{\alpha_t}\,x_{0,i}+\sqrt{1-\alpha_t}\,\epsilon_{t,i}, 
\end{equation}
where all dimensions evolve toward noise.

The decoder's predicted noise $e_\theta$, is subtracted from the current state $x_t$ to denoise all components incrementally. At the end of the reverse process, the class labels are rounded to the nearest integer to restore their discrete categorical nature.

\[
x_{t-1,i}\approx x_{t,i}-e_{\theta,i}.
\]

\noindent\textbf{Loss Function:}
Training minimizes the mean squared error between predicted and applied noise across all dimensions:
\begin{equation}
\text{MSE}=\tfrac{1}{n}\sum_{i=1}^n \|e_{\theta,i}-e_{\text{rand},i}\|^2. \label{eq:unguided_combined_loss}
\end{equation}
Unlike the guided case, no per-class Chamfer Distance is used since labels are diffused jointly with coordinates.

\subsection{Training Algorithm}

Training the model amounts to minimizing the defined loss functions, as outlined in Algorithms~\ref{alg:diffusion} and~\ref{alg:diffusion_un}. 

For the \textbf{guided} diffusion process, we optimize the combined losses in Eq.~\ref{eq:combined_loss}, ensuring reconstructions are both spatially accurate and semantically consistent. The spatial MSE penalizes errors in coordinate noise prediction, the label MSE enforces invariance in class channels, and the per-class CD preserves structure within each semantic category.  

For the \textbf{unguided} diffusion process, the objective reduces to the MSE across all point-cloud dimensions (Eq.~\ref{eq:unguided_combined_loss}), encouraging robust generation without semantic guidance.

\begin{algorithm}
\footnotesize
\caption{Guided 3D Diffusion Training Process}\label{alg:diffusion}
\begin{algorithmic}[1]
\State \textbf{Input:} $x_0 \in \mathbb{R}^{n \times 4}$ (initial point cloud), \texttt{encoder} (PointNet encoder), \texttt{var\_sched} (variance schedule)
\State \textbf{Output:} $loss$ (loss value)

\State $context \gets \texttt{encoder}(x_0)$
\State Initialize $t$ (time steps) if not provided
\State Calculate $\alpha\_bar \gets \texttt{var\_sched.alpha\_bars}[t]$
\State Calculate $\beta \gets \texttt{var\_sched.betas}[t]$
\State Compute $c_0 \gets \sqrt{\alpha\_bar}$ and $c_1 \gets \sqrt{1 - \alpha\_bar}$

\State Generate noise $e\_rand \in \mathbb{R}^{n \times 3}$ for spatial dimensions
\State $noised\_xyz \gets c_0 \cdot x_0[:, :3] + c_1 \cdot e\_rand$
\State $noised\_x \gets \texttt{concatenate}(noised\_xyz, x_0[:, 3:])$

\State $e_\theta \gets \texttt{diffusion\_net}(noised\_x, \beta, context)$

\State Compute  $L_{spatial} \gets \texttt{MSE}(e_\theta[:, :3], e\_rand)$
\State Compute $L_{label} \gets \texttt{MSE}(e_\theta[:, 3:], \mathbf{0})$

\State Reconstruct $recon\_xyz \gets noised\_xyz - e_\theta[:, :3]$
\State $recon \gets \texttt{concatenate}(recon\_xyz, x_0[:, 3:])$

\State Compute $L_{chamfer} \gets \texttt{ChamferDistance}(recon, x_0, x_0[:, 3])$

\State $loss \gets L_{spatial} + L_{label} + L_{chamfer}$

\State \Return $loss$
\end{algorithmic}
\end{algorithm}

\begin{algorithm}
\footnotesize
\caption{Unguided 3D Diffusion Training Process}\label{alg:diffusion_un}
\begin{algorithmic}[1]
\State \textbf{Input:} $x_0 \in \mathbb{R}^{n \times 4}$ 
\State \textbf{Output:} $loss$ (loss value)

\State $context \gets \texttt{encoder}(x_0)$
\State Initialize $t$ (time steps) if not provided
\State Calculate $\alpha\_bar \gets \texttt{var\_sched.alpha\_bars}[t]$
\State Calculate $\beta \gets \texttt{var\_sched.betas}[t]$
\State Compute $c_0 \gets \sqrt{\alpha\_bar}$ and $c_1 \gets \sqrt{1 - \alpha\_bar}$

\State Generate noise $e\_rand \in \mathbb{R}^{n \times 4}$ for all dimensions
\State $noised\_x \gets c_0 \cdot x_0 + c_1 \cdot e\_rand$

\State $e_\theta \gets \texttt{diffusion\_net}(noised\_x, \beta, context)$

\State Compute loss $L \gets \texttt{MSE}(e_\theta, e\_rand)$

\State \Return $L$
\end{algorithmic}
\end{algorithm}

\section{Experimental Setup}

\subsection{Dataset}

We used the ShapeNet-part dataset~\cite{mo2019partnet}, a subset of the ShapeNet dataset \cite{chang2015shapenet}, for our model evaluation. This dataset is a standard benchmark for 3D point cloud processing and is distinguished by its per-point part annotations. It consists of 24 object classes with hierarchical annotations at three levels of granularity, which is crucial for training and evaluating our model's ability to generate semantically coherent parts.

For synthesis and reconstruction experiments, we used the dataset’s predefined training and testing splits. However, recognizing the suboptimal split mentioned in \cite{yang2019pointflow}, we also evaluated our generation results on a randomized split to ensure a robust assessment of our model's performance.

\subsection{Implementation}

Following Luo~\cite{luo2021diffusion}, our implementation uses PointNet \cite{qi2017pointnet} as the latent shape encoder. We trained both the guided and unguided models for 100,000 batches with a batch size of 64. For evaluation, we used globalized CD as the primary metric for point cloud synthesis and reconstruction.

To evaluate the generative capability, we sampled from the latent shape encoding $z$ and a standard Gaussian cloud to iteratively perform the reverse diffusion process, yielding the generated point clouds. The synthesis experiments were designed to demonstrate the model's capacity to generate accurate point clouds, as the quality of the generated outputs is directly tied to the effectiveness of the learned latent encoding.

\section{Results and Discussion}

\begin{table*}[t]
    \centering
    \caption{Comparison of unguided pcd generation performance for different data splits. 
JSD, COV, and 1-NNA are multiplied by $10^{2}$, while MMD is multiplied by $10^3$.  
The best scores between data splits are highlighted in bold. $\uparrow$: the higher the better, $\downarrow$: the lower the better.}
    
\renewcommand{\arraystretch}{.85} 
{
    \begin{tabular}{ccc|cc|cc|cc}
        \hline
        \textbf{Category} & \multicolumn{2}{c}{\textbf{JSD ($\downarrow$)}} & \multicolumn{2}{c}{\textbf{MMD ($\downarrow$)}} & \multicolumn{2}{c}{\textbf{COV (\%, $\uparrow$)}} & \multicolumn{2}{c}{\textbf{1-NNA ($\downarrow$)}} \\
        \cmidrule(r){2-3} \cmidrule(r){4-5} \cmidrule(r){6-7} \cmidrule(r){8-9}
        & \textbf{preset} & \textbf{random} & \textbf{preset} & \textbf{random} & \textbf{preset} & \textbf{random} & \textbf{preset} & \textbf{random} \\
        \hline
        Bed-1 & 11.24 & \textbf{6.10} & 1.47 & \textbf{1.30} & 54.17 & \textbf{82.87} & 75.00 & \textbf{69.11} \\
        Chair-1 & \textbf{3.06} & 3.50 & 1.00 & \textbf{0.87} & \textbf{34.10} & 26.96 & \textbf{75.88} & 76.75 \\
        Chair-3 & 32.91 & \textbf{31.24} & 2.34 & \textbf{2.33} & 4.77 & \textbf{5.00} & 99.30 & \textbf{98.89} \\
        Dishwasher-3 & \textbf{2.87} & 8.447 & 0.74 & \textbf{0.68} & \textbf{89.82} & 65.71 & \textbf{60.53} & 77.21 \\
        Hat-1 & 11.35 & \textbf{7.40} & \textbf{0.60} & 0.81 & \textbf{100.00} & 78.61 & 68.75 & \textbf{57.26} \\
         Scissors-1 & 15.65 & \textbf{9.85} & 1.72 & \textbf{1.65} & 91.94 & \textbf{94.67} & \textbf{49.98} & 51.10 \\
        TrashCan-1 & 6.79 & \textbf{3.54} & 0.94 & \textbf{0.87} & 70.27 & \textbf{75.00} & 79.43 & \textbf{60.94} \\
        \hline
    \end{tabular}
}
\label{table:gen-results}
\end{table*}

\begin{table*}[t]
\centering
\caption{Comparison of AutoEncoder reconstruction error (Chamfer Distance multiplied by 10$^{2}$). Algorithm \textbf{U} and \textbf{G} refer to models trained with a semantically unguided and guided diffusion process, respectively. The number \textbf{1}, \textbf{2}, and \textbf{3} refer to the three levels of ShapeNet-Part segmentation:
coarse-, middle- and fine-grained. Dashed lines indicate levels that are not defined. The best scores between the unguided and guided models are highlighted in bold.}

\setlength{\tabcolsep}{2pt} 
\scalebox{0.87}
{
\begin{tabular}{l|r|rrrrrrrrrrrrrrrrrrrrrrrr}
\hline
     & \textbf{Avg$\downarrow$} & \textbf{Bag} & \textbf{Bed} & \textbf{Bott} & \textbf{Bowl} & \textbf{Chair} & \textbf{Clock} & \textbf{Dish} & \textbf{Disp} & \textbf{Door} & \textbf{Ear} & \textbf{Fauc} & \textbf{Hat} & \textbf{Key} & \textbf{Knife} & \textbf{Lamp} & \textbf{Lap} & \textbf{Micro} & \textbf{Mug} & \textbf{Frid} & \textbf{Scis} & \textbf{Stora} & \textbf{Table} & \textbf{Trash} & \textbf{Vase} \\ \hline
\textbf{U1 (Ours)} & 33.59 & 15.39 & 20.42 & 4.52 & 178.02 & 7.54 & 6.42 & 3.72 & 4.24 & 6.08 & \textbf{20.41} & 12.32 & 6.80 & 2.80 & \textbf{72.45} & 73.12 & 1.94 & 203.13 & \textbf{9.21} & 98.10 & 15.99 & 8.32 & 22.59 & 6.10 & 6.60 \\
\textbf{U2 (Ours)} & 50.40 & - & 44.23 & - & - & 50.17 & - & 5.88 & - & \textbf{7.15} & - & - & - & - & - & 168.51 & - & 7.90 & - & 7.79 & - & 25.29 & 136.67 & - & - \\
\textbf{U3 (Ours)} & 92.67 & - & 79.07 & 6.60 & - & 90.52 & 10.10 & 7.32 & 4.14 & 8.52 & 38.47 & 18.16 & - & - & \textbf{17.77} & 642.97 & - & 8.53 & - & 10.72 & - & 35.19 & 571.25 & 15.53 & 10.46 \\
\hline
\textbf{Avg} & 47.54 & 15.39 & 47.91 & 5.56 & 178.02 & 49.41 & 8.26 & 5.64 & 4.19 & 7.25 & 29.44 & 15.24 & 6.80 & 2.80 & \textbf{45.11} & 294.87 & 1.94 & 73.19 & \textbf{9.21} & 38.87 & 15.99 & 22.93 & 243.50 & 10.82 & 8.53 \\ \hline

\textbf{G1 (Ours)} & \textbf{15.54} & \textbf{14.83} & \textbf{16.16} & \textbf{4.09} & \textbf{44.90} & \textbf{6.62} & \textbf{5.24} & \textbf{2.84} & \textbf{4.14} & \textbf{5.73} & 20.56 & \textbf{6.48} & \textbf{6.16} & \textbf{2.33} & 85.31 & \textbf{12.55} & \textbf{1.84} & \textbf{64.16} & 10.98 & \textbf{23.59} & \textbf{6.89} & \textbf{6.06} & \textbf{11.51} & \textbf{4.69} & \textbf{5.18} \\

\textbf{G2 (Ours)} & \textbf{22.16} & - & \textbf{27.27} & - & - & \textbf{36.79} & - & \textbf{5.79} & - & 8.79 & - & - & - & - & - & \textbf{20.60} & - & \textbf{5.26} & - & \textbf{6.16} & - & \textbf{11.08} & \textbf{77.67} & - & - \\

\textbf{G3 (Ours)} & \textbf{20.38} & - & \textbf{49.89} & \textbf{4.30} & - & \textbf{43.05} & \textbf{8.26} & \textbf{6.04} & \textbf{3.68} & \textbf{6.88} & \textbf{33.49} & \textbf{7.03} & - & - & 23.65 & \textbf{21.59} & - & \textbf{6.21} & - & \textbf{6.62} & - & \textbf{11.22} & \textbf{101.97} & \textbf{6.92} & \textbf{5.66} \\
\hline
\textbf{Avg} & \textbf{19.36} & \textbf{14.83} & \textbf{31.11} & \textbf{4.20} & \textbf{44.90} & \textbf{28.82} & \textbf{6.75} & \textbf{4.89} & \textbf{3.91} & \textbf{7.13} & \textbf{27.03} & \textbf{6.76} & \textbf{6.16} & \textbf{2.33} & 54.48 & \textbf{18.25} & \textbf{1.84} & \textbf{25.21} & 10.98 & \textbf{12.12} & \textbf{6.89} & \textbf{9.45} & \textbf{63.72} & \textbf{5.81} & \textbf{5.42} \\
\hline

\end{tabular}
}
\label{tab:comparison}
\end{table*}

\subsection{Synthesis}

\begin{figure}[t]
    \centering
    \resizebox{0.95\columnwidth}{!}{%
        \begin{minipage}{\columnwidth}
            \begin{subfigure}[t]{\columnwidth}
                \centering
                \includegraphics[width=\textwidth]{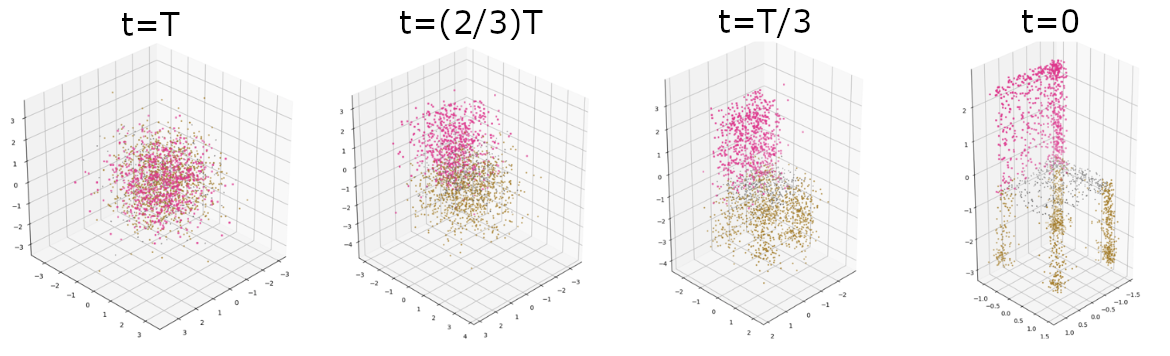} 
                \caption{Point cloud synthesis of a semantically annotated chair through conditional guided diffusion.}
                \label{fig:Guideddiffusion}
            \end{subfigure}

            \vspace{0.5em} 
            \begin{subfigure}[t]{\columnwidth}
                \centering
                \includegraphics[width=\textwidth]{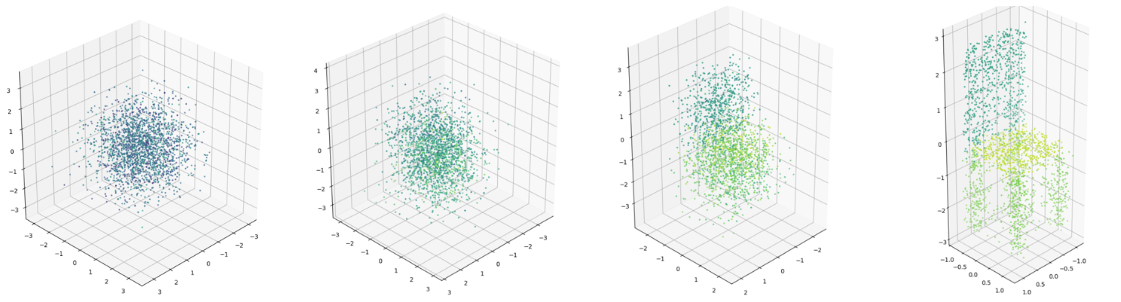}
                \caption{Point cloud synthesis of a semantically annotated chair through conditional unguided diffusion.}
                \label{fig:Unguideddiffusion}
            \end{subfigure}
        \end{minipage}
    }
    \caption{Examples of Guided and Unguided Diffusion process at different time steps. \textbf{T} represents the total diffusion time.}
    \label{fig:diffusion_graphics}
\end{figure}

The experimental results demonstrate the effectiveness of the guided diffusion process compared to the unguided diffusion process in synthesizing point clouds. Figures \ref{fig:Guideddiffusion} and \ref{fig:Unguideddiffusion} visually illustrate the point cloud synthesis through guided and unguided diffusion, respectively.

Table \ref{tab:comparison} presents a quantitative comparison of the AutoEncoder reconstruction error, measured by overall Chamfer Distance (multiplied by \(10^2\)). Lower Chamfer Distance (CD) values signify higher reconstruction quality. The table details the average reconstruction error for both the unguided and guided diffusion processes across various object categories and three levels of part-net segmentation.

The unguided process yielded average reconstruction errors ranging from 47.54 to 92.67, with finer segmentation levels showing higher errors. In contrast, the guided diffusion process significantly reduced the error to a range of 19.36 to 20.38, demonstrating its superior performance.

Analysis of these results highlights an interesting relationship between model performance, object complexity, and annotation fidelity. While some objects like microwaves and refrigerators benefited from increased annotation detail, others, such as chairs and tables, exhibited exponentially higher errors. Furthermore, categories with complex topologies, like lamps, consistently showed high reconstruction errors across all models and segmentation levels, suggesting a fundamental challenge in capturing their intricate structures within the latent space. 

These findings indicate that while guided diffusion generally improves performance, its effectiveness is highly dependent on the specific characteristics of the object and the level of annotation detail.

\subsection{Generation}

We evaluated our point cloud generation method using four metrics: Jensen-Shannon Divergence (JSD), Minimum Matching Distance (MMD), Coverage (COV), and 1-Nearest Neighbor Accuracy (1-NNA). The results, for both preset and random data splits, are summarized in Table \ref{table:gen-results} and visualized in Fig.\ref{fig:teaser}. Each metric offers a unique perspective on
the quality of the generated point clouds, with results presented
for both preset and random splits. JSD, based on the Kullback-Leibler divergence, measures the dissimilarity between the test and generated probability distributions. MMD evaluates the
overall quality of the generated point clouds. COV indicates the method’s ability to generate a diverse range of shapes and 1-NNA assesses how closely the test and generated sets
overlap. 

For point cloud generation, only the unguided model is evaluated. The guided method’s performance is heavily influenced by the hyperparameter class label ratios set during generation. For instance, generating a chair with the guided process requires setting the class labels by hand. If points only were given ’seat’ and ’legs’ labels, it may result in generating a stool. Therefore, evaluation metrics for the guided method are highly dependent on the chosen labels. To ensure a fair and consistent evaluation, we focused on the unguided method, which does not rely on specific hand-picked label ratios.

Random splits consistently outperformed preset splits across most metrics and categories, as exemplified by a category like Bed-1, where JSD improved from 13.43 to 6.41 and COV increased from 54.17\% to 81.58\%. This highlights the importance of a balanced data split for better generalization. The poor performance of certain categories, such as Chair-3, across both splits, suggests that the model struggles to effectively learn their underlying structure. These results show that random splits give a more reliable picture of the model’s generalization. However, the absolute values of the evaluation metrics should be interpreted in the context of the data distribution, since class imbalance or inherently complex categories can skew the scores.

\section{Conclusion}
We introduced a diffusion-based framework for 3D point cloud generation that integrates semantics directly into the generative process. Prior methods either omit semantics or impose them post hoc through external segmentation or clustering, decoupling part-level understanding from geometry during generation. Our approach produces point clouds that are structurally coherent and part-aware, and our analysis of guided and unguided diffusion shows that semantic conditioning yields faithful, interpretable, and semantically consistent generations. By treating semantics as an integral generative signal, this work points to a promising direction for point cloud diffusion models and provides a basis for future research on controllable, semantically grounded 3D generation.

\bibliographystyle{IEEEtran}
\bibliography{refs}

\end{document}